\documentclass[10pt,twocolumn,letterpaper]{article}
\usepackage{cvpr}              % To produce the CAMERA-READY version
\usepackage{amsmath}

\definecolor{cvprblue}{rgb}{0.21,0.49,0.74}
\usepackage[pagebackref,breaklinks,colorlinks,allcolors=cvprblue]{hyperref}
\usepackage{amsmath}
\usepackage{algorithm}
\usepackage{algorithmic}
\usepackage{tabularx}
\usepackage{caption}
\usepackage{float}
\usepackage{multirow}

\usepackage[most]{tcolorbox} % 绘制框
\usepackage{graphicx}       % 插入图片
\usepackage{xcolor}         % 定义颜色
\def\paperID{11018} % *** Enter the Paper ID here
\def\confName{CVPR}
\def\confYear{2025}

\usepackage{xcolor}
\title{\textit{STEP}: Enhancing Video-LLMs' Compositional Reasoning by \\ \underline{S}patio-\underline{Te}mporal Gra\underline{p}h-guided Self-Training}
\author{Haiyi Qiu$^{1*}$ \quad Minghe Gao$^{1*}$ \quad  Long Qian$^1$ \quad Kaihang Pan$^1$ \quad Qifan Yu$^1$ \quad Juncheng Li$^{1\dagger}$\\
Wenjie Wang$^2$ \quad Siliang Tang$^1$ \quad Yueting Zhuang$^1$ \quad Tat-Seng Chua$^2$ \\
$^1$Zhejiang University \qquad $^2$ National University of Singapore
}

\begin{document}

\maketitle
\begin{abstract}

Video Large Language Models (Video-LLMs) have recently shown strong performance in basic video understanding tasks, such as captioning and coarse-grained question answering, \textbf{but} struggle with compositional reasoning that requires multi-step spatio-temporal inference across object relations, interactions, and events. The hurdles to enhancing this capability include extensive manual labor, the lack of spatio-temporal compositionality in existing training data and the absence of explicit reasoning supervision. In this paper, we propose \textbf{STEP}, a novel graph-guided self-training method that enables Video-LLMs to generate reasoning-rich fine-tuning data from any raw videos to improve itself. Specifically, we first induce Spatio-Temporal Scene Graph (STSG) representation of diverse videos to capture fine-grained, multi-granular video semantics. Then, the STSGs guide the derivation of multi-step reasoning Question-Answer (QA) data with Chain-of-Thought (CoT) rationales. Both answers and rationales are integrated as training objective, aiming to enhance model's reasoning abilities by supervision over explicit reasoning steps. Experimental results demonstrate the effectiveness of \textbf{STEP} across models of varying scales, with a significant 21.3\% improvement in tasks requiring three or more reasoning steps. Furthermore, it achieves superior performance with a minimal amount of self-generated rationale-enriched training samples in both compositional reasoning and comprehensive understanding benchmarks, highlighting the broad applicability and vast potential. 
\makeatletter{\renewcommand*{\@makefnmark}{}
\footnotetext{$^*$Equal contribution.
\textsuperscript{$^\dagger$}Corresponding author.}
}
\end{abstract}    
\vspace{-1em}
\section{Introduction}
\label{sec:intro}

\begin{figure}[t]
  \centering
   \includegraphics[width=\linewidth]{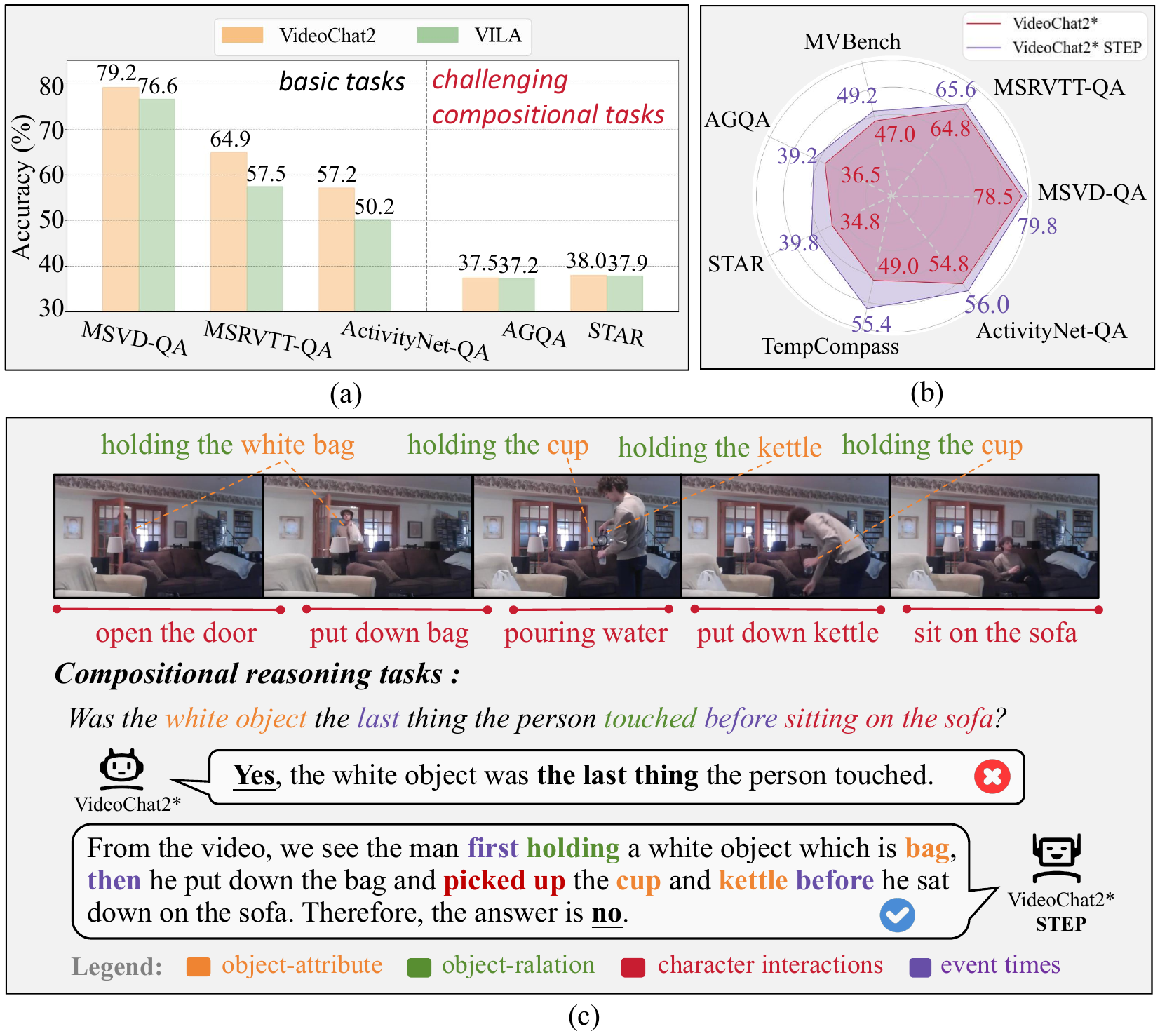}
   \vspace{-2.2em}
   \caption{(a) Top left: A significant performance gap between standard understanding and compositional reasoning tasks for advanced Video-LLMs. (b) Top right: Notable improvement with our method. (c) Bottom: An example illustrating the challenging tasks and our performance gains.}
   \label{fig:fig1}
   \vspace{-1.5em}
\end{figure}

Recently, Video Large Language Models (Video-LLMs) such as VideoChat \cite{li2023videochat}, Video-LLaMA \cite{zhang2023video}, and Video-LLaVA \cite{lin2023video} have demonstrated impressive results in the field of video understanding,  particularly in global interpretive tasks like video captioning, coarse-grained visual question answering, and general summarization \cite{li2024mvbench,ataallah2024minigpt4,liu2024st,maaz2023video}. 
% gmh % 
However, recent empirical studies \cite{fei2024video,mitra2024compositional} show that even the most advanced Video-LLMs struggle with the compositional reasoning tasks that require multi-step spatio-temporal reasoning across diverse object attributes, relations, dynamic character interactions and events, as shown by a significant performance gap in Figure \ref{fig:fig1} (a). Compositional reasoning is essential to understand complex visual semantics of open-world videos \cite{kurby2008segmentation,lillo2014discriminative,reynolds2007computational,speer2007human}, while its absence hinders Video-LLMs from advancing toward real-world applications, as shown in the example in Figure \ref{fig:fig1} (c).

Several studies \cite{fei2024video,mitra2024compositional,chen2024grounded} have attempted to address the challenge, but notable limitations remain:

\textbf{1) Extensive manual labor and lack of generalization}: Although compositional datasets such as CLEVRER \cite{yi2019clevrer}, TVQA \cite{lei2018tvqa}, and NExT-QA \cite{xiao2021next} have been developed as fine-tuning resources to enhance models' reasoning abilities \cite{li2024mvbench}, the human-annotated data construction demands substantial manual effort, making it impractical to generate large-scale training samples. Moreover, methods relying solely on those datasets are task-specific and often lack the flexibility to generalize to new, unseen scenarios. \textbf{2) Inadequacy of spatio-temporal compositionality}: Video semantics are typically extracted using limited clip-level descriptors \cite{yu2023anetqa, grunde2021agqa, wu2024star}, which restrict the richness of visual interactions and temporal dynamics, thereby hindering a deeper understanding of spatio-temporal details in videos \cite{li2022fine}. Additionally, large-scale datasets generated through prompting LLMs \cite{achiam2023gpt} tend to yield simplistic questions, limiting the training of models to decompose complex problems and perform multi-step reasoning. \textbf{3) Absence of explicit supervision for reasoning process}: Current black-box training methods compute only the loss between model output and ground truth \cite{wei2021finetuned, zhou2023instruction}, causing models to rely on spurious correlations \cite{mitra2023orca} instead of structured intermediate 
reasoning steps (“rationales”) behind answers. This lack of supervision hinders the ability of compositional reasoning, where multiple reasoning steps need to be well combined in a coherent sequence. How to effectively and controllably obtain multi-step rationales to guide this reasoning process remains an open question \cite{li2024vocot, gao2024fact}. \textbf{In summary}, an ideal learning paradigm would not only generate compositional training data enriched with multi-granular spatio-temporal video details, but also provide explicit reasoning supervision to better train Video-LLMs.

In this paper, we propose a novel graph-guided video self-training method: \textbf{STEP}, enabling the model to self-generate fine-grained and reasoning-rich fine-tuning data from any raw videos to improve itself. Specifically, \textbf{1)} we perform the \textbf{symbolic structure induction} of Spatio-Temporal Scene Graph (STSG) from any raw videos by four defined operations: visual splitting, semantics parsing, dynamic merging, and cross-clip bridging, to capture multi-granular and fine-grained video semantics, enabling a structured representation of spatial and temporal details in video. \textbf{2)} We implement a \textbf{stepwise graph-driven rationale learning} process on the structured STSG representations, sampling multi-step reasoning paths to generate diverse, reasoning-rich Question-Answer (QA) tasks along with step-by-step Chain-of-Thought (CoT) rationales. Then we train the model to learn both the answers and the rationales as integral components of the training objective, distilling the reasoning process to enhance its capability for complex, multi-step compositional reasoning.

In our framework, we take advantage of Video-LLMs’ capability for self-training, greatly reducing reliance on extensive human-annotated data. By employing the STSG as a unified structured foundation to encapsulate complex video semantics, the model effectively captures fine-grained spatial relationships and temporal dynamics with high fidelity, enhancing the framework’s capacity to generate compositional tasks across multiple video hierarchies. Moreover, our stepwise graph-driven rationale learning process allows the model to draw from the inherent reasoning logic within the graph structure, aligning each step in the rationale precisely with sub-questions in compositional tasks. By incorporating these well-reasoned, interpretable rationales as integral components of the training objective, we significantly enhance the model’s compositional reasoning abilities.

Extensive experiments show that \textbf{STEP} notably enhances the compositional reasoning performance of Video-LLMs with different parameters and architectures, especially with a 21.3\% improvement on tasks requiring three or more reasoning steps. Furthermore, compared to models trained on manually annotated datasets, \textbf{STEP} achieves superior model performance across diverse benchmarks, with a minimal amount of self-generated, reasoning-rich training samples, highlighting the broad applicability and vast potential.
Our contributions can be summarized as follows:

\begin{itemize}
    \item We introduce \textbf{STEP}, a novel graph-guided self-training method that leverages spatio-temporal scene graphs to guide the model in self-generating reasoning-rich QA tasks and CoT rationales for training, thereby enhancing its compositional reasoning abilities.
    \item \textbf{STEP} is model-agnostic, enabling easy application across various Video-LLM architectures, and is designed to operate with minimal manual effort, effectively leveraging large-scale raw unlabeled videos for training.
    \item With a smaller dataset size, \textbf{STEP} shows improved performance not only on complex compositional reasoning datasets, but also on standard VQA, comprehensive and long video understanding benchmarks, underscoring the effectiveness and vast potential of our approach.

\end{itemize}

%-------------------------------------------------------------------------
\section{Related Work}
\label{sec:related}
\textbf{Video Large Language Models (Video-LLMs)}. Following the notable success of Large Language Models (LLMs) \cite{li2023llama, chowdhery2023palm,openai2023chatgpt}, many works have adapted LLMs to the video modality \cite{achiam2023gpt,li2023videochat,lin2023video,qian2024momentor}, aiming to combine LLMs' reasoning and interactive skills with video perception. These methods align visual features with LLMs' feature space via projection layers, enabling tasks like video captioning and QA. However, current Video-LLMs remain at the perceptual surface of videos, lacking fine-grained spatio-temporal understanding and compositional reasoning abilities.

A notable effort, Video-of-Thought (VoT) \cite{fei2024video}, integrates STSG representations into the model input for pixel-level spatio-temporal understanding and applies CoT prompts for step-to-step task decomposition. However, it needs specialized training for STSG encoder, adding computational overhead, and relies on custom CoT prompts for specific tasks, limiting generalization and scalability. In contrast, our approach is more versatile to apply across various Video-LLM architectures. It requires no additional modules to encode STSG representation, instead extracting rich semantics in STSG into fine-grained QA and reasoning-rich rationales, enhancing adaptability across various reasoning tasks.

\noindent \textbf{Visual Instruction Tuning and Self-Training.}
Numerous works \cite{hua2024v2xum,liu2024improved,zhang2024videoinstructiontuningsynthetic,li2023fine} have demonstrated the importance of visual instruction tuning for improving Video-LLMs’ performance. However, the high cost and inefficiency of manual annotation hinder large-scale data collection for compositional reasoning. Consequently, self-training methods \cite{zelikman2022star,huang2022large,amini2022self}, where LLMs autonomously generate training data, have gained traction for scalable instruction tuning.

Video-STAR \cite{zohar2024video}, as the first video self-training approach, has shown the method’s feasibility. However, it relies on labeled metadata, limiting the scope of available datasets, and uses simplistic prompts for generating training data, leading to lower-quality training data for complex reasoning tasks. Our method, by contrast, requires no manual annotation and can directly process raw, untrimmed videos. By leveraging STSG representation, it captures fine-grained spatio-temporal details, enhancing compositional reasoning while offering a more reasoning-rich training data.

\section{Method}
\label{sec:method}
\begin{figure*}[t]
    \centering
    \includegraphics[width=\textwidth]{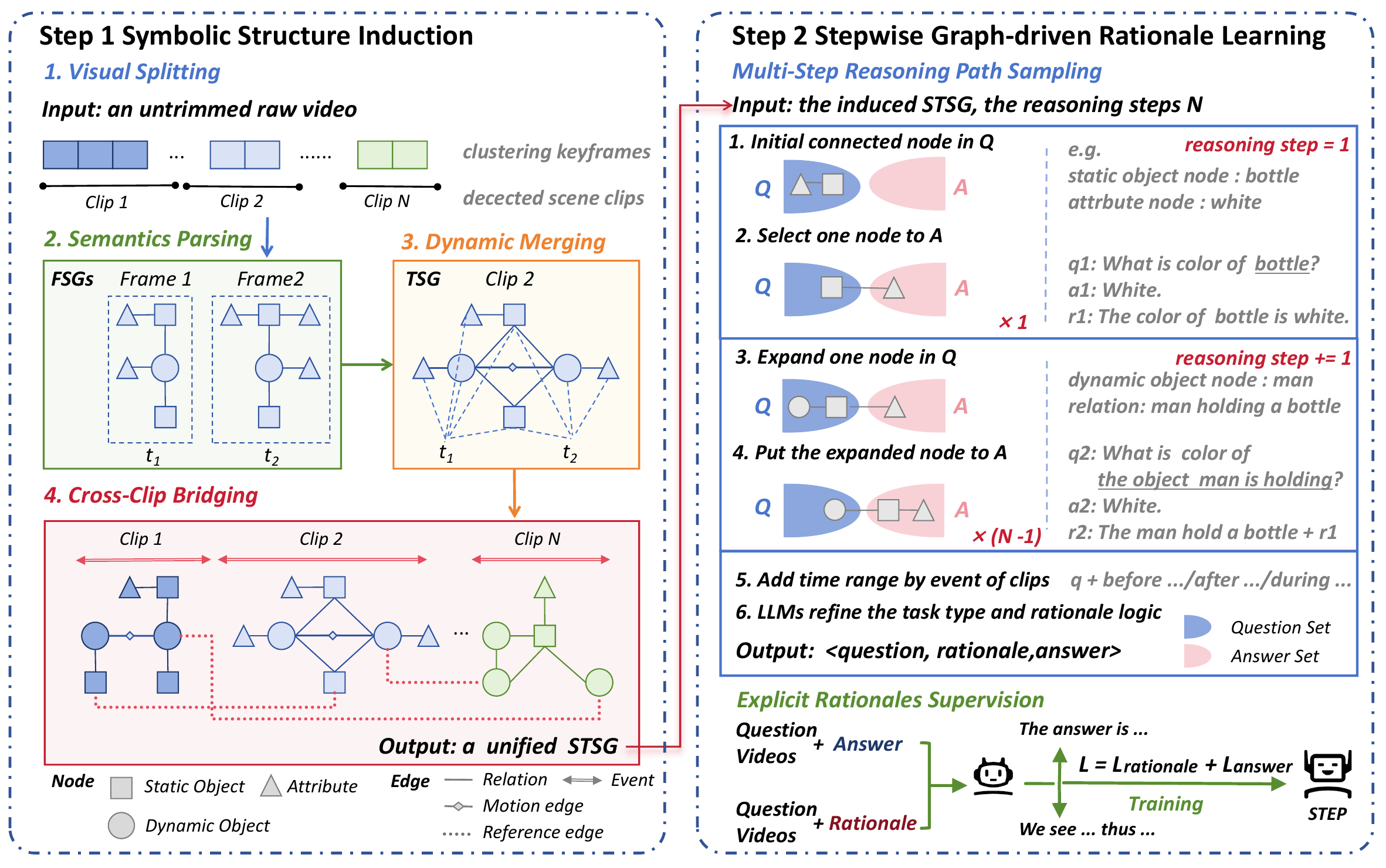}
    \vspace{-2em}
    \caption{\textbf{A high-level overview of our STEP approach.} We first perform symbolic structure induction to convert spatio-temporal details into a unified STSG. Then a graph-driven rationale learning process is implemented to generate QA pairs with CoT rationales from reasoning paths, providing explicit supervision during training.}
        
    \label{fig:fig2}
    \vspace{-1.2em}
\end{figure*}
To enhance compositional reasoning in Video-LLMs with minimal manual effort, we introduce \textbf{STEP}, a model-agnostic graph-guided self-training method allowing Video-LLMs to effectively generate reasoning-rich training data for improving itself, as depicted in Figure \ref{fig:fig2}. Given a raw video, we first perform symbolic structure induction to abstract the intricate visual content into a structured STSG representation (Section \ref{sec:3.1}). 
We then implement a stepwise graph-driven rationale learning process to derive QA pairs with CoT rationales from reasoning paths on STSGs, providing explicit supervision during training (Section \ref{sec:3.2}). 

\subsection{Symbolic Structure Induction}
\label{sec:3.1}
Raw videos are saturated with chaotic, unstructured and redundant visual information, making direct utilization for model training challenging. While prior work \cite{herzig2019spatio,ji2020action,nguyen2024hig} has shown the effectiveness of structured video representations, it is primarily centered on object-level semantics and constrained by rule-based extraction, missing fine-grained spatio-temporal details. Inspired by \cite{yu2023visually}, we design a systematic paradigm to induce the model to symbolize raw videos into a unified, open-vocabulary and fine-grained STSG. Four defined operations — visual splitting, semantics parsing, dynamic merging, and cross-clip bridging — effectively capture and organize multi-granular spatio-temporal details into the nodes and edges of the STSG, encompassing objects, relations, actions, and events, thereby enabling more structured and comprehensive reasoning.

\noindent \textbf{Visual Splitting}. Given an untrimmed raw video, we use PySceneDetect \cite{castellano2018pyscenedetect} to detect scene cuts and segment them into distinct clips, capturing various scene transitions. Then a clustering-based extraction method \cite{song2016click} is applied to obtain representative keyframes, so as to maintain fine-grained key semantics while minimizing redundant features.

\noindent \textbf{Semantics Parsing}. For each keyframe at time $t
$, we design a series of purpose-driven parsing instructions to guide the model to automatically generate Frame Scene Graph (FSG), denoted as $ G_t = (O_t, A_t, R_t) $. 
%(see Appendix  \textcolor{red}{A.1} for the details).
More specifically, we induce a set of \textbf{object nodes}  $ O_t = \{o_1, o_2, \dots, o_n\} $ from scene narrative of the keyframe, then instruct the model to categorize them into static or dynamic. For each object $ o_{i} $, we request a detailed description to extract its fine-grained\textbf{ attribute nodes}, contributing to the set of attribute nodes $ A_t = \{a_{i,j} \mid o_i \in O_t\} $. Subsequently, for each pair of objects $(o_i, o_j)$, we construct subject-predicate-object triples to capture their relational correspondence, forming \textbf{relation edges} $ r_{i,j} = (o_i, p_{i,j}, o_j) $, where $ p_{i,j} $ describes their relationship. These edges collectively define the set $ R_t = \{r_{i,j} \mid o_i, o_j \in O_t\} $. To reduce potential hallucinations and inaccuracies, we employ a dual verification process: \textit{(i)} sampling $n$ responses to compute node/edge frequencies as confidence scores and discarding low-confidence ones; \textit{(ii)} prompting the model to verify each node/edge’s presence in the video, discarding those labeled as “no.” This ensures reliable visual information extraction. 

\noindent \textbf{Dynamic Merging}. While FSGs capture fine-grained visual semantics, the short temporal intervals between consecutive frames often introduce redundant nodes and edges, hindering computation and propagation \cite{wu2020graph}.
To address it, we merge identical static object nodes across frames into a unified node, preserving essential attributes and updating the connected edges to maintain spatial relationships. 
For dynamic nodes, we introduce \textbf{motion edges} $m_k = (o_{i,t_1}, p_k, o_{i,t_2};[t1,t2])$ to succinctly capture the motion relationship, where $o_{i,t_1}$ and $o_{i,t_2}$ denote the same object $o_i$ at different timestamps, $p_k$ describes the motion type, and $[t_1, t_2]$ specifies the temporal interval over which this motion occurs. The set $M_k = \{m_k\}$ allows the model to capture and differentiate object movements over time, reducing redundancy while enhancing the representation of dynamic interactions. The resulting graph, termed a Temporal Scene Graph (TSG), integrates static and dynamic elements, providing a rich foundation for temporal reasoning tasks requiring analysis of object trajectories and interactions..

\noindent \textbf{Cross-clip Bridging}. While TSGs provide comprehensive intra-clip spatial and temporal information, cross-clip relations remain underrepresented. 
To bridge it, we introduce \textbf{reference edges}  between object nodes across clips, ensuring semantic coherence and temporal continuity. To determine if an object $o_{i}$ in clip $c_1$ corresponds to an object $o_{j}$ in clip $c_2$, we input their respective keyframes, along with extracted labels and attributes, into the Video-LLM, prompting it to assess whether the specified objects are identical. This enhances the model's ability to consistently track objects across scenes, thereby supporting tasks that require long-term temporal reasoning and continuity. Additionally, we obtain \textbf{event edges} for each clip, providing a holistic description and view of all clips. 

Ultimately, we extract fine-grained visual information at the frame level, merge redundant details, integrate dynamic motions, and bridge cross-clip relation information, resulting in a unified STSG representation.
\subsection{Stepwise Graph-driven Rationale Learning}
\label{sec:3.2}
The induced STSGs represent the spatio-temporal structure of videos, providing a wealth of fine-grained visual details and  dynamic interactions for compositional learning. However, the intricate nature of these graph structures renders it impractical to directly apprehend and integrate them into the reasoning mechanisms of models, whether as inputs or outputs \cite{yun2024compositional,li2023variational}. 
Motivated by the insight that reasoning tasks can be generated from a structured hierarchical graph \cite{jin2024graph,zhou2024enhancing,li2021adaptive}, we propose a multi-step reasoning path sampling method to compose visual semantics of nodes and edges into structured compositional question-answer, while simultaneously producing step-to-step CoT rationales which reflect an explicit reasoning process for graph-inferable answers. Finally, We implement explicit rationales supervision, where both the answers and their corresponding rationales are integrated into the training objective, thereby enhancing model's compositional reasoning.

\noindent \textbf{Multi-step Reasoning Path Sampling}. Considering that each node on STSG represents a visual semantic  in videos, any pair of connected nodes can form a single-step visual question. To facilitate the construction of intricate multi-step reasoning tasks, we sample diverse reasoning paths that traverse multiple nodes and edges across the graph. The length of each path, corresponding to the number of reasoning steps, enables precise control over task complexity, allowing for a balanced integration of straightforward queries and advanced multi-step reasoning challenges.

\begin{figure}[t]
  \centering
   \includegraphics[width=\linewidth]{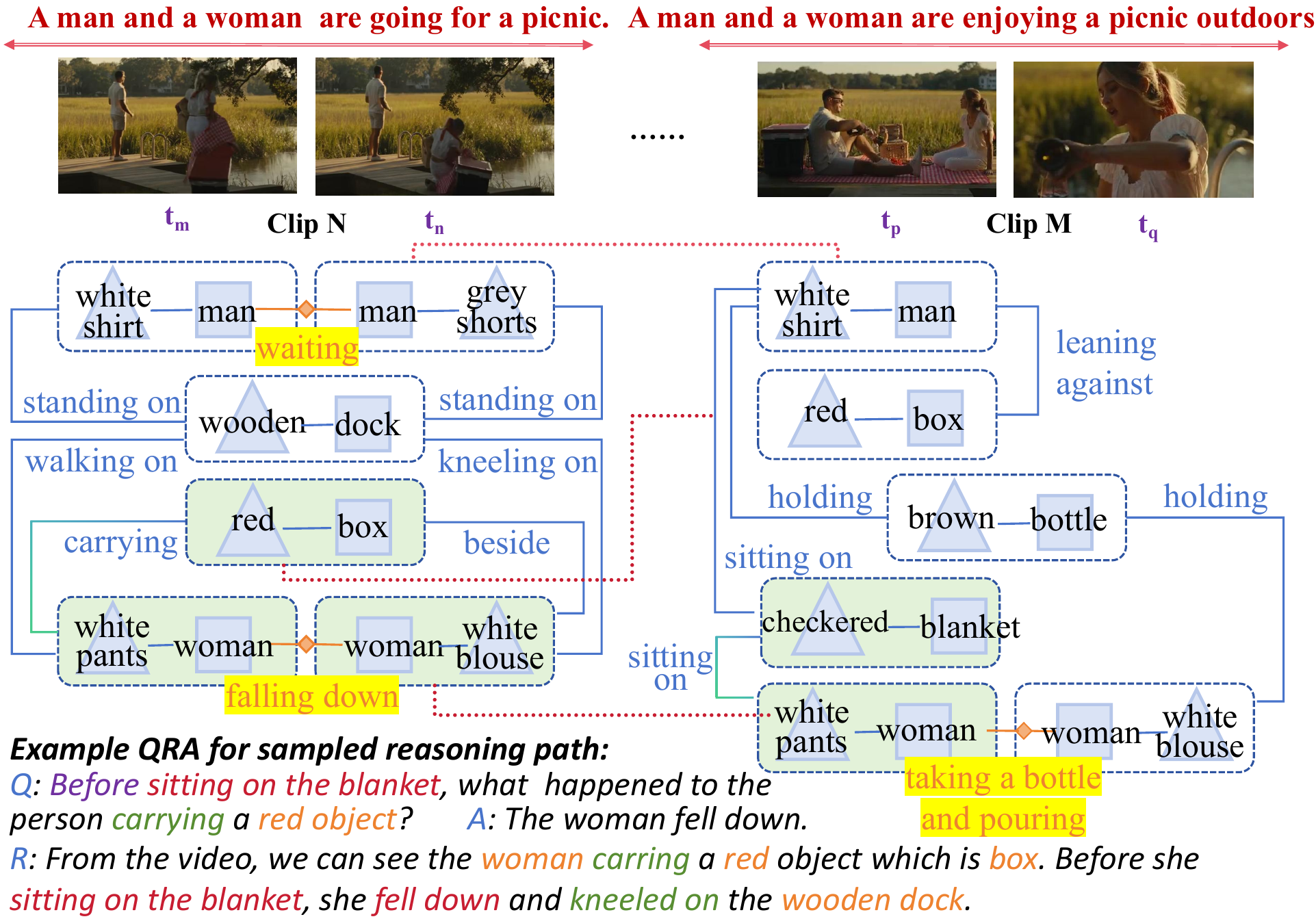}
   \vspace{-1.5em}
   \caption{Examples of the process that construct STSG and generate Question-Rationale-Answer (QRA) samples}
  \vspace{-2em}
   \label{fig:fig3}

\end{figure}
Given the spatio-temporal scene graph $ G $ and a specified number of reasoning steps $ N \in \mathbb{Z}^{+} $, we iteratively sample a reasoning path $ p $ by expanding it over $ N $ iterations. 
\textbf{1) Initialization}: we begin with an empty path $ p_0 $ and initialize two sets: a question set $ Q = \emptyset $ and an answer set $ A = \emptyset $. The nodes in $ Q $ correspond to the components of the current question, indicating the parts of the reasoning that remain open for expansion. In contrast, the nodes in $ A $ have already been incorporated as answers to previous sub-questions and are no longer expandable.
\textbf{2) N-step Expansion}: we first randomly select a  pair of connected nodes from $ G $, one in $ Q $ and another in $ A $, generating the initial question with answer. In each subsequent iteration, we randomly select a node from  $ Q $ and expanding its connected nodes, progressively transforming the question into a more complex form and adding one more step to the reasoning path. The expanded node is then moved to $ A $, indicating that it has been fully expanded. This process continues until no further nodes can be expanded from $ Q $, or the maximum number of reasoning steps $ N $ is reached. \textbf{3) Temporal Contextualization}: to incorporate temporal aspects, we select an event edge and apply a time range to the question, thus grounding the question in a specific temporal context. 

In this process, each node expansion corresponds to the addition of a new sub-question, representing a discrete reasoning step within the multi-step inference process. As each expansion, we record the corresponding sub-question and answer, progressively building a richer and more detailed CoT rationale. Finally, we obtain not only the complex, multi-step question with answer but also the explicit CoT rationale that outlines how this answer was derived through the series of reasoning steps.
We then utilize the language model within the Video-LLMs to diversify the QA types, enhance the logical flow in the rationales (see Appendix \textcolor{red}{A.2}  for more details), thereby enabling tasks to be unconstrained by templates, more diverse and adaptable.

\setlength{\tabcolsep}{4pt} 

\begin{table*}[t]
\centering
\renewcommand{\arraystretch}{0.9}
\begin{tabularx}{\textwidth}{@{}l *{6}{c} | *{4}{c} @{}}
\toprule
& \multicolumn{6}{c}{\textbf{Zero-shot Standard QA Datasets}} & \multicolumn{4}{c}{\textbf{Compositional Reasoning Datasets}} \\

\cmidrule(lr){2-7} \cmidrule(lr){8-11}
 & \multicolumn{2}{c}{MSVD-QA} & \multicolumn{2}{c}{MSRVTT-QA} & \multicolumn{2}{c|}{ActivityNet-QA} & \multicolumn{2}{c}{AGQA} & \multicolumn{2}{c}{STAR} \\
& Accuracy & Score & Accuracy & Score & Accuracy & Score  & Accuracy & Score & Accuracy & Score \\
\midrule
VideoChat (7B) & 56.3 & 2.8 & 45.0 & 2.5 & 26.5 & 2.2 & - & - & - & - \\
VideoChatGPT (7B) & 64.9 & 3.3 & 49.3 & 2.8 & 35.2 & 2.7 & - & - & - & -\\
Video-LLaVA (7B) & 70.7 & 3.9 & 59.2 & 3.5 & 45.3 & 3.3 & 34.8 & 2.8 & 24.9 & 2.6 \\
VideoChat2 (7B)  & 79.2 & 4.0 & 64.9 & 3.4 & 57.2 & 3.5 & 37.5 & 2.9 & 38.0 & 2.7 \\
\midrule
VideoChat2* & 78.5 & 3.9 & 64.8 & 3.4 & 54.8 & 3.4 & 36.5 & 2.9 & 34.8 & 2.5 \\
VideoChat2* $Instruct$ & 78.9 & 3.9 & 64.1 & 3.3 & 55.0 & 3.4 & 37.4 & 2.9 & 36.2 & 2.5 \\
VideoChat2* $Distillation$ & 79.0 & 3.9 & 65.0 & 3.4 & 55.2 & 3.4 & 38.2 & 3.0 & 37.6 & 2.6 \\
VideoChat2* \textbf{STEP} & \textbf{79.8} & \textbf{4.0} & \textbf{65.6} & \textbf{3.5} & \textbf{56.0} & \textbf{3.5} & \textbf{39.2} & \textbf{3.2} & \textbf{39.8} & \textbf{2.8} \\
\midrule
VILA (3B) & 76.6 & - &  57.5 & - & 50.2 & - &  37.3 & 3.1 & 37.9 & 2.7  \\
VILA $Instruct$ & 77.0 & 3.8 &  56.5 & 3.2 & 53.3 & 3.4 &  37.4 & 3.1 & 38.0 & 2.7  \\
VILA $Distillation$ & 77.2 & 3.8 &  58.9 & 3.3 & 51.4 & 3.3 &  38.3 & 3.1 & 39.4 & 2.8  \\
VILA \textbf{STEP} & \textbf{78.2} & \textbf{3.9} & \textbf{60.6} & \textbf{3.3} & \textbf{55.1} & \textbf{3.5} & \textbf{38.9} & \textbf{3.2} & \textbf{40.3} & \textbf{2.8} \\
\bottomrule
\end{tabularx}
\vspace{-0.5em}
\captionsetup{font=normalsize} 
\caption{Comparison of model performance on zero-shot standarad QA and compositional reasoning datasets}
\vspace{-1em}
\label{tab:table1}
\end{table*}

\noindent \textbf{Explicit Rationales Supervision}.
To address the lack of explicit supervision over the model's intermediate reasoning steps, which is inherent in traditional black-box training, we incorporate generated rationales into the training process. These rationales are not merely supplementary inputs, but play a crucial role by providing transparency into the model's reasoning at each step. Rather than treating the rationales as isolated components, we frame the learning process as a multi-task problem, where both the answers and their corresponding rationales are jointly learned to enhance the model's reasoning ability. In other words, the $f(x,q,i^{a})\to\hat{a}$ and $f(x,q,i^r)\to\hat{r}$ are trained with:
\vspace{-0.5em}
\begin{align}
L_{\text{answer}} &= \frac{1}{N} \sum_{k=1}^N l(f(x_k, q_k, i^a_k), \hat{a}_k) \tag{1} \\
L_{\text{rationale}} &= \frac{1}{N} \sum_{k=1}^N l(f(x_k, q_k, i^r_k), \hat{r}_k) \tag{2}
\end{align}

\vspace{-0.5em} The $\hat{a}$ denotes the answer to the compositional question $q$ of video $x$, and $\hat{r}$ represents the corresponding CoT rationales. Here, $i^a$ and $i^r$ are distinct instructions for answer and rationale generation, respectively. This formulation enables the model to predict task answers while internalizing the reasoning process. The loss function is defined as:
\vspace{-0.5em}
\begin{equation}
L=L_{\text{answer}}+\lambda L_{\text{rationale}} \tag{3}
\end{equation}

\vspace{-0.5em} We set $\lambda$ to 1 to guarantee equal priority for answer prediction and rationale generation. This equilibrium in our approach highlights our dedication to fostering a model that is adept at not only producing accurate predictions but also articulating coherent and logical rationales.
\section{Experiments}
\label{sec:experiments}

\textbf{STEP} is a model-agnostic method that bootstraps compositional reasoning QA pairs with step-by-step CoT rationales for effectively self-training Video-LLMs. In this section, we outline our experimental setup (Section \ref{sec:4.1}), evaluate against several baselines on various compositional reasoning and video understanding tasks (Section \ref{sec:4.2}), and assess model performance across different reasoning steps (Section \ref{sec:4.3}). We also conduct ablation studies to investigate the contributions of each operation in STSG generation, the impact of $\lambda$ in loss function and the impact of reasoning steps on rationales for training (Section \ref{sec:4.4}).

\subsection{Experimental Setup}
\label{sec:4.1}

\textbf{Model Setup}. We compare \textbf{STEP} against two models with different parameter sizes as backbones: VideoChat2 with Mistral 7B \cite{li2024mvbench} and VILA 3B \cite{lin2024vila},  aiming to show that our method is model-agnostic and effective across architectures. 

\noindent \textbf{Initial Model}.
Most self-training frameworks start with a pre-trained model to generate more detailed explanations from labeled datasets \cite{zohar2024video,zelikman2024star,fang2024vila}. However, our framework is designed to operate on any unlabeled raw videos, requiring the initial model to have a baseline level of instruction-following capability to carry out the complex STSG construction tasks involved. To meet this requirement, we first perform instruction tuning on the pre-trained (visual-language aligned) VideoChat2 model using a small set of existing instruction-tuning data (see Appendix \textcolor{red}{B} for details), resulting in the baseline model, VideoChat2*. For VILA, as pre-trained checkpoints are unavailable, we conduct experiments directly on its instruction-tuned model.

\noindent \textbf{Traning Settings}. We train the initial model using our proposed framework, along with two control models:
\begin{itemize}
    \item The \textbf{STEP} model employs explicit rationales supervision on self-generated QRA training data to demonstrate the framework’s effectiveness. 
    \item The \textit{Instruct} model is trained on an existing manually annotated dataset to compare performance with our smaller but rationale-rich dataset. 
\item The \textit{Distillation} model leverages GPT-4V \cite{openai2023gpt4} to generate QRA training samples for training, providing a basis for comparing self-training against model distillation. 
\end{itemize}

1) For the \textbf{STEP} model, we employ the \textbf{STEP} framework to guide the backbone in autonomously generating reasoning-rich QRA training samples from raw video datasets and fine-tuning itself using the loss functions in Section \ref{sec:3.2}. The process is iterative to better leverage the model’s enhanced capabilities in each cycle,  enabling progressive improvement in reasoning through repeated data generation and training, ultimately forming a self-enhancing closed-loop mechanism. 
2) The \textit{Instruct} model is trained on a manually annotated dataset derived from the same raw videos used by \textbf{STEP} and doubled in size. Since these traditional datasets only contain questions and answers, the training loss $L = L_{\text{answer}}$.
3) The \textit{Distillation} model utilizes a stronger model GPT-4V to replace the self-training mechanism employed in \textbf{STEP} and use the same raw video sources, training loss function, and dataset size as in \textbf{STEP}, enabling a direct comparison of the effects of self-training versus model distillation on performance.

\noindent \textbf{Baselines.}
The paper also lists results from other Video-LLMs like 
mPLUG-Owl \cite{ye2023mplug}, VideoChat \cite{li2023videochat}, VideoChatGPT \cite{achiam2023gpt}, Video-LLaVA \cite{lin2023video} for comparison.

\noindent \textbf{Evaluation Details.} We evaluate our method using the following benchmarks: \textbf{1)} compositional reasoning benchmarks, including AGQA \cite{grunde2021agqa} and STAR \cite{wu2024star}, by converting the source datasets into open-ended questions and applying the evaluation protocol from \cite{achiam2023gpt}. This protocol reports two metrics: accuracy (the percentage of correctly answered questions) and the average score (where ChatGPT rates each response on a scale of 0-5, with the mean score calculated). \textbf{2)} Zero-shot standard QA datasets, including MSVD-QA \cite{xu2017video}, MSRVTT-QA \cite{xu2016msr}, and ActivityNet-QA \cite{caba2015activitynet}, evaluated using the same protocol as 1). \textbf{3)} Comprehensive video understanding benchmarks, such as MVBench \cite{li2024mvbench} and TempCompass \cite{liu2024tempcompass}, following their respective evaluation methodologies. \textbf{4)} Long video understanding benchmarks, such as MovieChat-1K \cite{song2024moviechat} and MLVU \cite{zhou2024mlvu}, adhering to their evaluation protocols. All evaluations are conducted using the same GPT model \cite{wu2024freeva} (“gpt-3.5-turbo”) to ensure consistent comparisons across all tasks. We present the evaluation details in Appendix \textcolor{red}{D}.

\setlength{\tabcolsep}{9pt}
\begin{table*}[t]
\centering
\renewcommand{\arraystretch}{0.9}
\begin{tabularx}{\textwidth}{@{}lcccccccc}
\toprule
 & \multicolumn{4}{c}{\textbf{AGQA}} & \multicolumn{4}{c}{\textbf{STAR}} \\ 
  \cmidrule(lr){2-5} \cmidrule(lr){6-9} 
   & 1-step & 2-step & \(\geq\)3-step & All & 1-step & 2 step & \(\geq\) 3-step & All \\ 
\midrule
Step Distribution & 21.65 & 45.30 & 33.05 & 100 & 24.57 & 56.04 & 19.39 & 100 \\
\midrule
VideoChat2 (7B)*& 56.2 & 34.1 & 27.0 & 36.5 &  44.0 & 33.1 & 27.7 & 34.8\\ 
VideoChat2* \textit{Instruct} & 57.5 & 34.6 & 27.4 &  37.4 &46.9& 34.3 & 28.0 & 36.2 \\ 
VideoChat2* \textit{Distillation} & 57.9 & 35.4 & 29.2 & 38.2  & 46.9 & 35.5 & 31.8 & 37.6 \\ 
VideoChat2* \textbf{STEP} & \textbf{58.5} & \textbf{36.7} & \textbf{30.0 (11.1\%)} & \textbf{39.2} &  \textbf{47.7 }& \textbf{38.4} & \textbf{33.6 (21.3\%)}& \textbf{39.8 }\\ 
\midrule
VILA (3B) & 57.5 & 34.7 & 27.7 & 37.3  & 46.7 & 37.1& 29.1 &  37.9 \\ 
VILA \textit{Instruct} & 57.6 & 34.8 & 27.7 & 37.4 &  47.0 &  37.1 & 29.0 & 38.0    \\ 
VILA \textit{Distillation} & 58.2 & 35.7 & 28.8 & 38.3 & 47.0 & 38.5 & 32.1  & 39.4 \\ 
VILA \textbf{STEP} & \textbf{58.6} & \textbf{36.1} & \textbf{29.7 (7.2\%)} &  \textbf{38.9} &  \textbf{47.8}& \textbf{39.3} & \textbf{32.4 (11.3\%)}& \textbf{40.3}\\ 
\bottomrule
\end{tabularx}
\vspace{-0.5em}
\captionsetup{font=normalsize} 
\caption{Performance evaluation of compositional reasoning tasks over various reasoning steps on AGQA and STAR datasets.}
\vspace{-1em}
\label{tab:table2}
\end{table*}
\setlength{\tabcolsep}{2.1pt}
\begin{table}[b]
\vspace{-1.5em}
\centering
\resizebox{\linewidth}{!}{
\begin{tabularx}{1.3\linewidth}{@{}l@{}cccc}
\toprule
& \multicolumn{2}{c}{\textbf{Comprehensive}} & \multicolumn{2}{c}{\textbf{Long-video }} \\
\cmidrule(lr){2-3} \cmidrule(lr){4-5}
& TempCompass & MVBench & MovieChat-1K & MLVU \\
VideoChatGPT (7B) & 35.2 & 32.7 & 47.6 & 31.3\\
mPLUG-Owl-V (7B) &  40.0 & 29.7 & - & 25.9 \\
\midrule
VideoChat2 (7B)* & 49.0 & 47.0 & 63.5 & 43.5 \\
VideoChat2* $Instruct$ & 51.8 & 47.9 & 64.5 & 44.3\\
VideoChat2* $Distillation$ & 53.6 & 48.5 & 66.1 & 46.0\\
VideoChat2* \textbf{STEP} & \textbf{55.4} & \textbf{49.2} & \textbf{67.6} & \textbf{46.4}\\
\midrule
VILA (3B) & 51.4 & 43.0 & 55.4& 22.7\\
VILA $Instruct$ & 52.6 & 44.2 & 56.3 & 22.9 \\
VILA $Distillation$ & 53.5 & 44.8  & 56.9& 23.5\\
VILA \textbf{STEP} & \textbf{54.4} & \textbf{45.6} & \textbf{57.4}& \textbf{24.3}\\
\bottomrule 
\end{tabularx}
}
% \captionsetup{font=normalsize} 
\caption{Comparison of TempCompass, MVBench, MovieChat-1K and MLVU benchmark. For TempCompass, we present the results for the Multi-Choice QA task type. See more evaluation result details in Appendix \textcolor{red}{E}.}
\label{tab:table3}
\end{table}
\subsection{Quantitative Results}
\label{sec:4.2}
Results can be seen in Table \ref{tab:table1}. The notable performance gap between baseline models on standard QA and compositional reasoning tasks highlights the necessity of \textbf{STEP} for improving reasoning abilities. Moreover, \textbf{STEP} achieves significant performance enhancement on two backbones with varying architectures and parameter sizes, demonstrating the model-agnostic nature and effectiveness.

\noindent \textbf{Advanced performance on diverse video reasoning and understanding task}. As demonstrated by the AGQA and STAR datasets, \textbf{STEP} outperforms the baseline in compositional reasoning tasks, highlighting the effectiveness of the graph-guided self-training method in enhancing the model's reasoning capabilities. Moreover, improvements on standard QA datasets indicate that \textbf{STEP} extends beyond reasoning tasks, showing strong generalization and adaptability to a wide range of general video understanding tasks. 

\noindent \textbf{Explicit utility of rationales}. Compared to the \textit{Instruct} model, which uses twice the amount of manually annotated instruction-tuning data relative to our QRA samples, our method still achieves significantly greater improvements in reasoning tasks. This suggests that the reasoning-rich training data generated by \textbf{STEP} offer more effective support for enhancing the model’s reasoning capabilities than traditional datasets. Furthermore, the incorporation of explicit rationale supervision during training facilitates more effective internalization of reasoning, offering an advantage over conventional black-box training methods.

\begin{figure}[h]
  \centering

\includegraphics[width=0.8\linewidth]{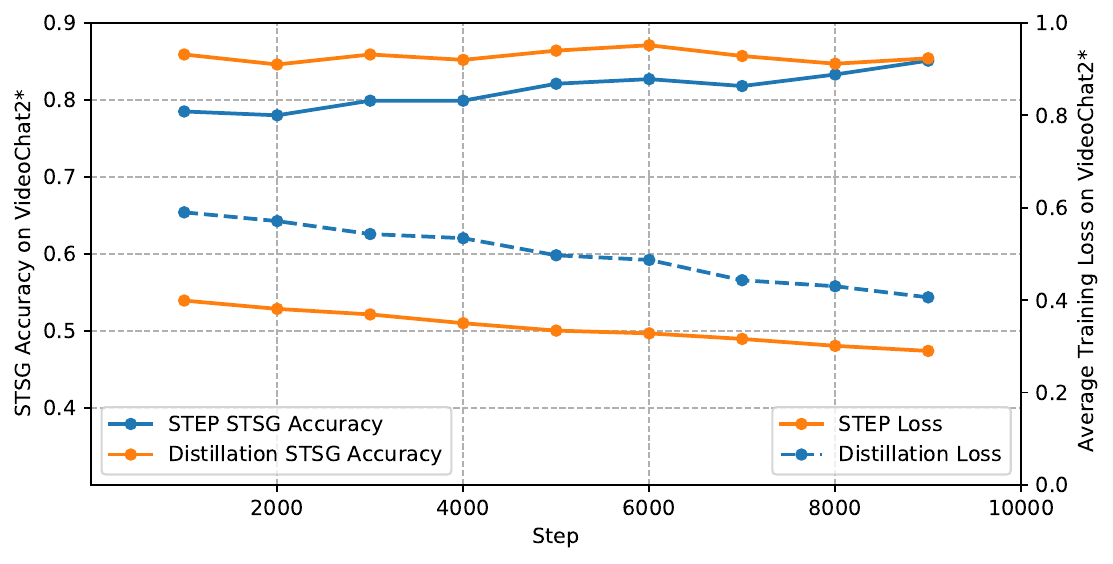}
   \vspace{-1em}
   \caption{The generated STSG accuracy (see measured details in Appendix \textcolor{red}{D}) and training loss of STEP and Distillation.}
   \vspace{-1em}
   \label{fig:fig4}
\end{figure}

\noindent \textbf{Superiority of self-training}.
Our self-training framework \textbf{STEP} outperforms \textit{Distillation} despite using a relatively weaker base model. By comparing generated STSG accuracy and training loss in Figure \ref{fig:fig4},  we attribute \textbf{STEP}’s superiority to: 1) \textbf{STEP} maintains comparable STSG accuracy through filtering, guaranteeing relatively precise generated QRAs. 2) Teacher model effectiveness significantly depends on compatibility with base models \cite{xu2024stronger}, \textbf{STEP}’s lower loss indicates better alignment with the base model’s knowledge and capabilities.
3) Unlike Distillation’s single-pass generation, STEP progressively produces STSGs and QRAs with improved abilities, fostering a positive feedback loop that enhances data quality and overall performance.

\noindent \textbf{Improvement on challenging benchmarks}. We evaluate \textbf{STEP} on four challenging benchmarks, showing significant improvements (Table \ref{tab:table3}). 
TempCompass and MVBench are fine-grained temporal benchmarks sensitive to hallucinations, showing that our method effectively interprets event sequences and reduces hallucinations by integrating multi-granular details and query-aligned reasoning steps. MovieChat-1K and MLVU are  minute-long video benchmarks with diverse content, demonstrating robust generalization and enhanced long-video understanding in models. 

\subsection{Performance Analysis Over Reasoning Steps}
\label{sec:4.3}
We further measure the model performance across different reasoning steps to better understand its reasoning capabilities, as shown in Table \ref{tab:table2}. For the AGQA dataset, we utilize the reasoning steps provided, which are based on the “ground-truth” reasoning path derived from its scene graph. For STAR questions, since no related data is available, we manually assigned a number of reasoning steps to each question template to standardize evaluation.

Notably, we observe that, aligned with the overall performance, our \textbf{STEP} approach outperforms the baseline and control models across all reasoning step cases in both compositional datasets. We attribute this improvement to the advantages conferred by reasoning-rich training data and explicit rationale-based supervision during training. Particularly, on the challenging reasoning tasks requiring three or more reasoning steps, our method achieves a remarkable improvement of 21.3\% in STAR datasets, highlighting its effectiveness  for enhancing complex reasoning abilities.

\subsection{Ablation Study}
\label{sec:4.4}
\textbf{Qualitative analysis}.
We present a qualitative example in Figure \ref{fig:fig3} to illustrate our process from an untrimmed raw video to a unified STSG representation, which finally becomes reasoning-rich QRA training samples. We also show an example of improved model performance in Figure \ref{fig:fig4}.

\begin{figure}[t]
  \centering
   \includegraphics[width=\linewidth]{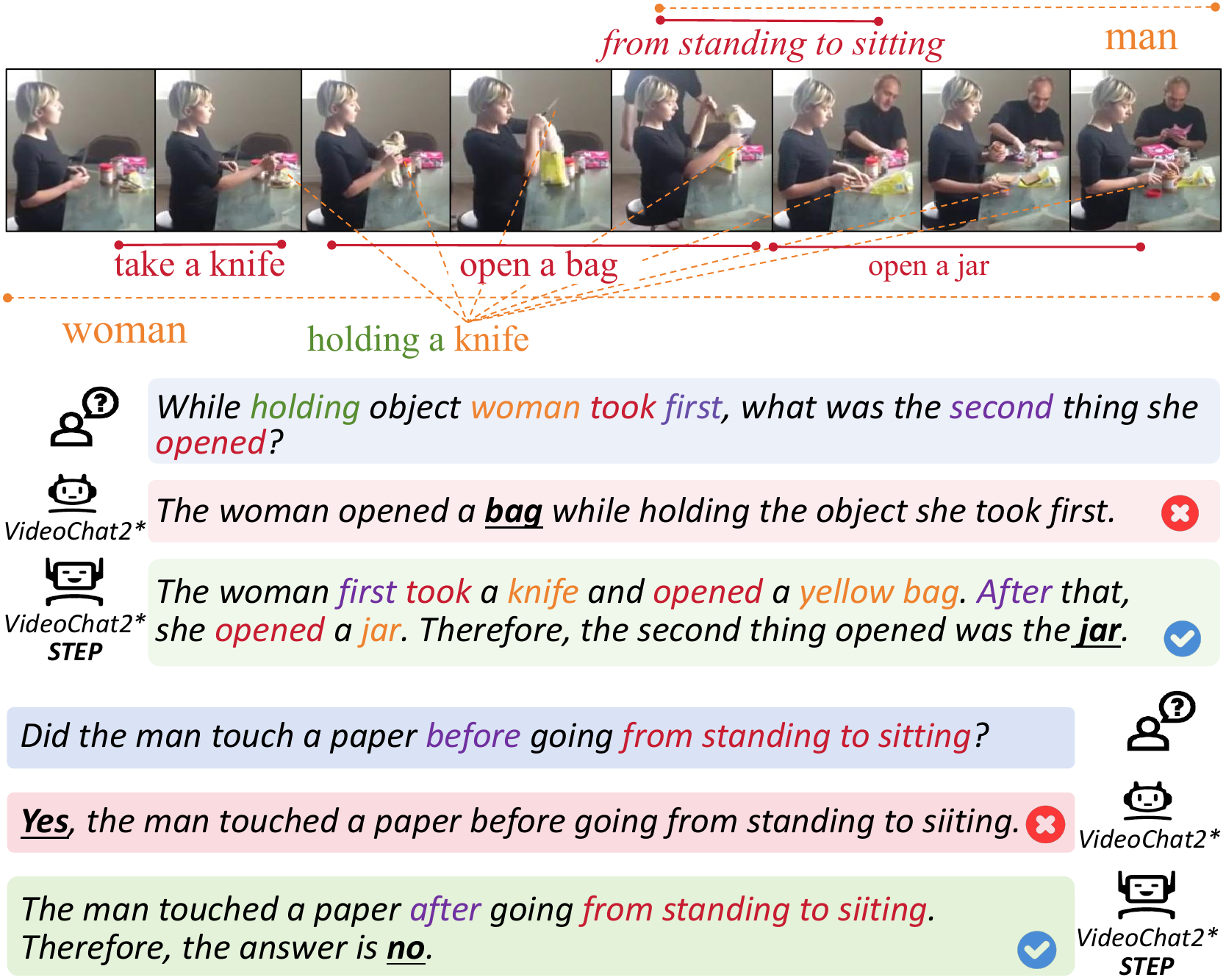}
   \vspace{-1.5em}
   \caption{Example outputs of the model trained by STEP}
  \vspace{-1.5em}
   \label{fig:fig5}

\end{figure}

\setlength{\tabcolsep}{3.4pt}
\begin{table}[b]
\vspace{-1.5em}
\centering
\small
\begin{tabularx}{0.40\textwidth}{lcccc}
\toprule
&  \multicolumn{2}{c}{AGQA} & \multicolumn{2}{c}{STAR} \\
% \cmidrule(lr){2-3} \cmidrule(lr){4-5} 
& Accuracy & Score & Accuracy & Score \\
\midrule
1 \hspace{0.1em} \textbf{STEP} & \textbf{39.2} & \textbf{3.2} & \textbf{39.8} & \textbf{2.8} \\
2 \hspace{0.5em}  w/o splitting & 38.9 & 3.2 & 39.2 & 2.8 \\
3 \hspace{0.5em}  w/o parsing & 37.8 & 3.0  & 36.5 & 2.6\\
4 \hspace{0.5em}  w/o merging & 38.3 & 3.1 & 37.3 & 2.7 \\
5 \hspace{0.5em}  w/o bridging & 38.6 & 3.1 & 38.3 & 2.7 \\

\bottomrule
\end{tabularx}

% \captionsetup{font=normalsize} 
\caption{Ablation results (\%) of individual components.}
\label{tab:table4}
\end{table}

\noindent \textbf{Analysis on STSG induction}. 
For the four operations of STSG induction, we conduct sequential ablation experiments using VideoChat2* as the backbone, with results on AGQA and STAR shown in Table ~\ref{tab:table4}: 1) \textbf{STEP}: applies all operations to construct a unified STSG for generating QRA samples; 2) \textbf{STEP} w/o splitting: divides videos into uniform time intervals and samples frames evenly, rather than detecting scene transitions and extracting key frames via clustering; 3) \textbf{STEP} w/o parsing: directly employs a simplified prompt to generate a JSON-format scene graph without incremental extraction of attributes, objects and relations; 4) \textbf{STEP} w/o merging: leaves redundant nodes and dynamic information unprocessed; and 5) \textbf{STEP} w/o bridging: omits cross-clip information. We observe that parsing is the most crucial operation, enabling extraction of multi-granular spatial-temporal details. Merging and bridging are also essential for reducing redundancy and preserving dynamic information, while splitting has the least impact, as uniform time intervals still provide sufficient structure.

\begin{figure}[h]
  \centering
  \begin{subfigure}[b]{0.48\linewidth}
    \centering
    \vspace{-1em}
    \includegraphics[width=\linewidth]{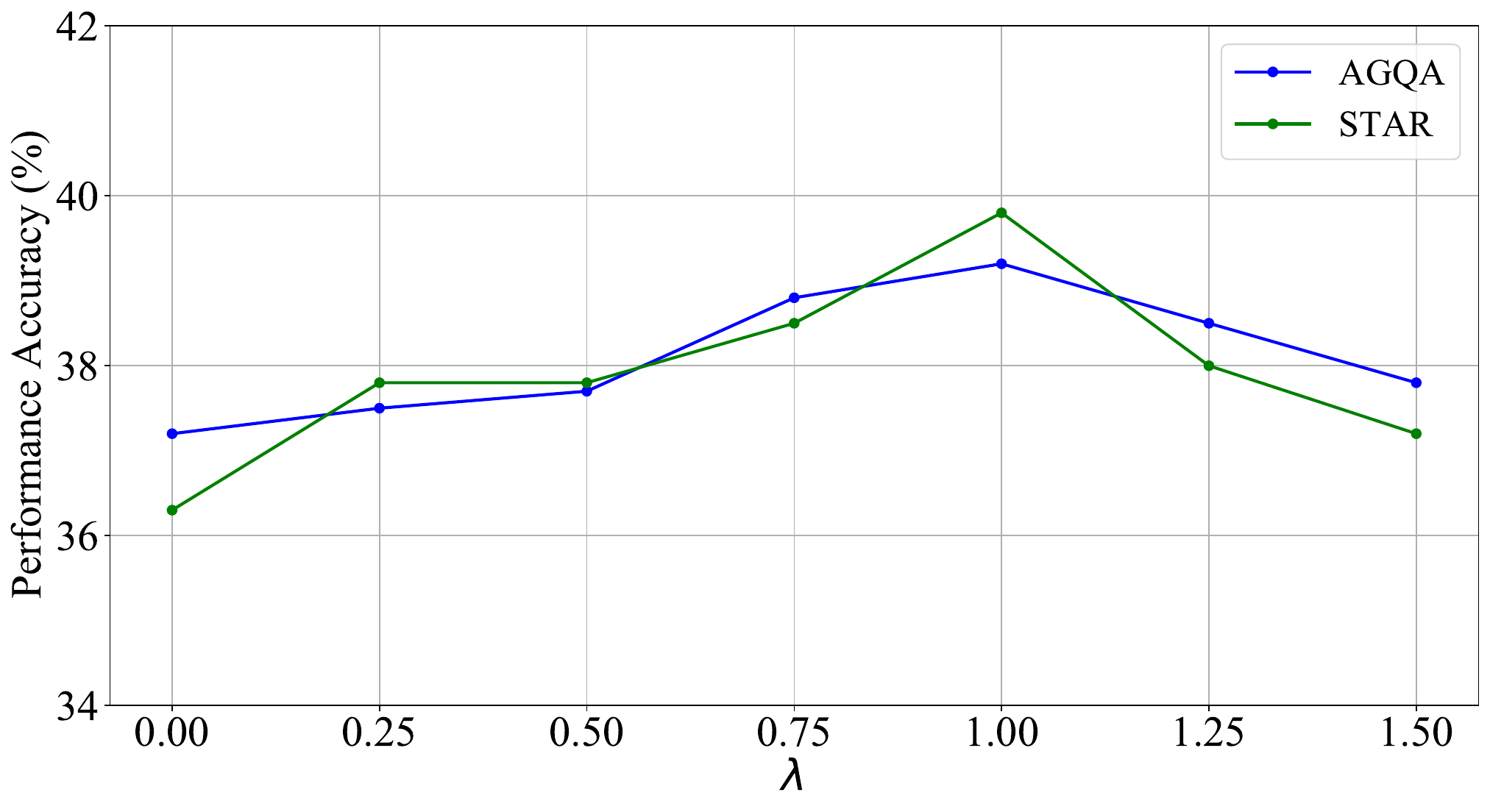}
    \vspace{-1.2em}
    \caption{}
    \label{fig:fig5a}
  \end{subfigure}
  \hfill
  \begin{subfigure}[b]{0.48\linewidth}
    \centering
    \includegraphics[width=\linewidth]{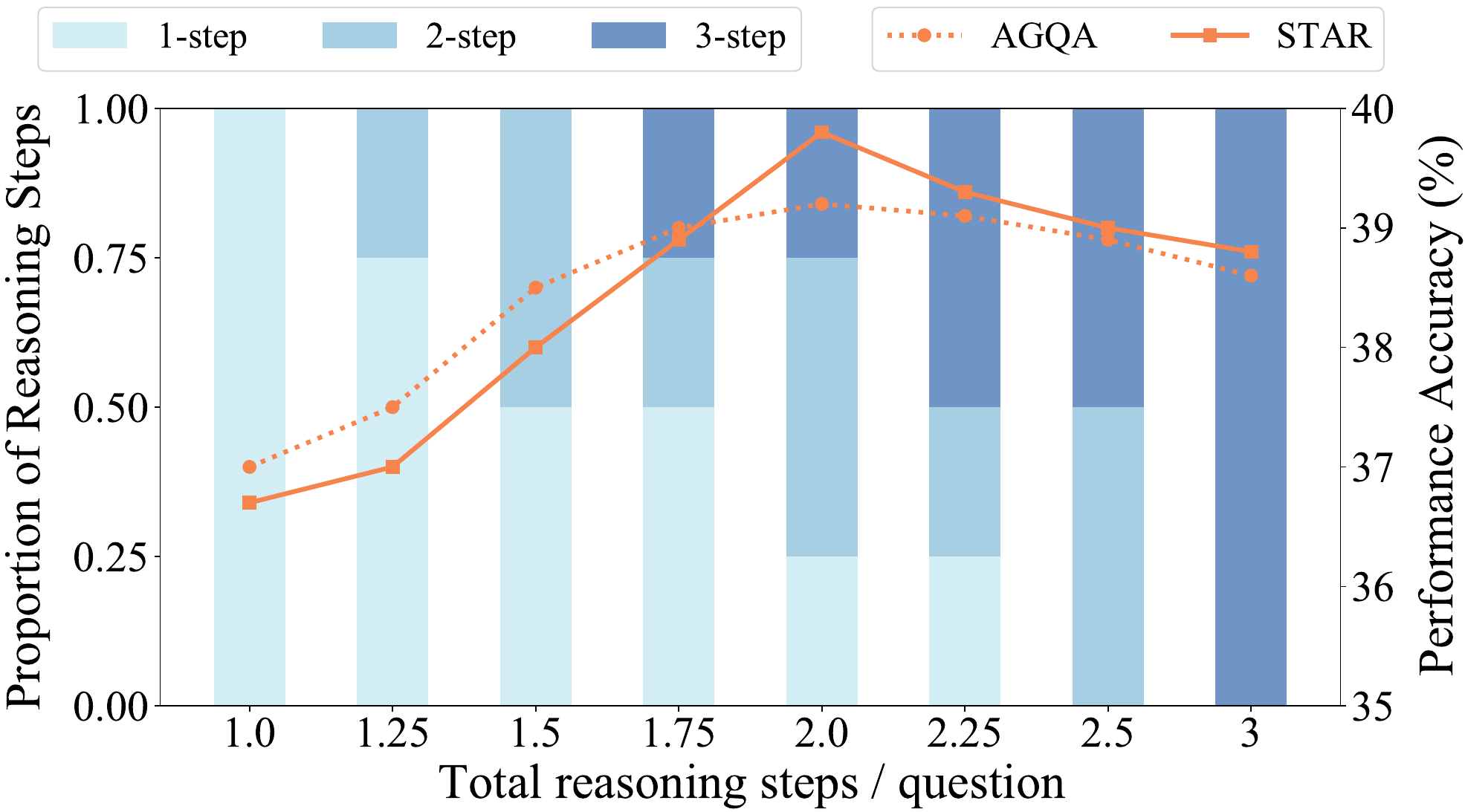}
    \vspace{-1.2em}
    \caption{}
    \label{fig:fig5b}
  \end{subfigure}
  \vspace{-0.8em}
  \caption{(a) Impact of $\lambda$ on model performance.  (b) Impact of reasoning step distributions}
  \label{fig:combined}
  \vspace{-1em}
\end{figure}

\noindent \textbf{Analysis on $\lambda$ in rationale supervision.}
We explore the impact of $\lambda$ in the loss function (Section \ref{sec:3.2}) on the trade-off between answer accuracy and rationale quality, as illustrated in Figure \ref{fig:fig5a} on VideoChat2*. The results indicate that $\lambda = 1$ achieves the best performance, as smaller values fail to sufficiently train rationale reasoning, while larger values cause the rationale's intricacy to dominate, thereby negatively impacting answer accuracy.

\noindent \textbf{Analysis on reasoning steps of rationales.}
We examine the effect of reasoning step distributions in QRA training samples on model performance using VideoChat2* as the backbone in Figure \ref{fig:fig5b}. We find that an overabundance of simple 1-step questions leads to performance degradation, likely due to limited reasoning exposure. Conversely, using only complex 3-step questions also reduces performance, suggesting that overly complex samples hinder generalization. The best results are achieved with a balanced distribution of reasoning steps, allowing the model to learn from both simple and complex samples.

\section{Conclusion}
\label{sec:conclusion}
% In conclusion, we propose \textbf{STEP}, a model-agnostic graph-guided self-training method that leverages STSG representation to self-generate fine-grained, reasoning-rich training data from any raw videos, employing a stepwise explicit rationale learning process to boost Video-LLMs' capabilities on multi-step reasoning tasks. Extensive experiments show that \textbf{STEP} improves compositional reasoning by 21.3\% on complex multi-step tasks and outperforms models trained on annotated datasets, with minimal self-generated, reasoning-rich training samples. It achieves strong performance on comprehensive and long video understanding benchmarks on two different backbones, demonstrating its broad applicability and potential in advancing reasoning in Video-LLMs.

In conclusion, we introduce \textbf{STEP}, a model-agnostic, graph-guided self-training framework that utilizes STSG representations to  self-generate fine-grained, reasoning-rich training data from raw videos. By incorporating a stepwise explicit rationale learning mechanism, \textbf{STEP} significantly enhances the multi-step reasoning capabilities of Video-LLMs. Extensive experimental results demonstrate that \textbf{STEP} achieves a 21.3\% improvement in compositional reasoning on complex multi-step tasks, surpassing models trained on manually annotated datasets, even with a minimal amount of self-generated, reasoning-rich training samples. Furthermore, \textbf{STEP} exhibits robust performance on comprehensive and long-video understanding benchmarks across two distinct backbones, underscoring its broad applicability and potential to advance reasoning in Video-LLMs.

\noindent \textbf{Acknowledgement.} This work was supported by the NSFC (62272411), the Key R\&D Projects in Zhejiang Province (No. 2024C01106, 2025C01030), the Zhejiang NSF (LRG25F020001). 
\newpage

{
    \small
    \bibliographystyle{ieeenat_fullname}
    \bibliography{main}
}


% \input{X_suppl}
% {
%     \small
%     \bibliographystyle{ieeenat_fullname}
%     \bibliography{main}
% }
% \input{rebuttal}
\end{document}